\documentclass{article}
\usepackage{spconf,amsmath,graphicx}

\usepackage{lipsum}
\usepackage{amssymb}
\usepackage{siunitx}
\usepackage{algorithmic}
\usepackage{commath}
\usepackage{amssymb}
\usepackage{color}
\usepackage{threeparttable}
\usepackage{url}
\usepackage{booktabs}
\usepackage{tabularx}
\usepackage{media9}
\usepackage{mathtools}
\usepackage{subcaption}
\usepackage{multicol}

\title{Learning to Segment Corneal Tissue Interfaces in OCT Images}

\name{Tejas Sudharshan Mathai $^{1\star}$, Kira L Lathrop $^{2,3}$, and John Galeotti $^{1,2}$}
\address{$^{1}$ The Robotics Institute, Carnegie Mellon University, USA	\\
$^{2}$ Department of Bioengineering, University of Pittsburgh, USA	\\
$^{3}$ Department of Ophthalmology, University of Pittsburgh, USA}

\begin{document}


\maketitle

\begin{abstract}
Accurate and repeatable delineation of corneal tissue interfaces is necessary for surgical planning during anterior segment interventions, such as Keratoplasty. Designing an approach to identify interfaces, which generalizes to datasets acquired from different Optical Coherence Tomographic (OCT) scanners, is paramount. In this paper, we present a Convolutional Neural Network (CNN) based framework called CorNet that can accurately segment three corneal interfaces across datasets obtained with different scan settings from different OCT scanners. Extensive validation of the approach was conducted across all imaged datasets. To the best of our knowledge, this is the first deep learning based approach to segment both anterior and posterior corneal tissue interfaces. Our errors are 2$\times$ lower than non-proprietary state-of-the-art corneal tissue interface segmentation algorithms, which include image analysis-based and deep learning approaches.
\end{abstract}
\section{Introduction}
\label{sec:intro}

Optical Coherence Tomography (OCT) is an imaging modality used to visualize corneal \cite{Izatt1994}, limbal \cite{Kira2012}, and retinal structures \cite{Huang1991} with micrometer resolution. OCT can be used to estimate corneal biometric parameters \cite{Kuo2012}, such as corneal curvature and refractive power, and it has been integrated into surgical microscopes for use in surgical procedures such as cataract surgery, LASIK, and Deep Anterior Lamellar Keratoplasty (DALK) \cite{Kuo2012,Keller2018}. Accurate reconstruction of the cornea and estimation of these parameters for clincal use requires precise delineation of corneal tissue interfaces, thereby aiding surgeons with their surgical planning.

\begin{figure}[h]
\centering
\begin{subfigure}[b]{0.125\columnwidth}
\centering
\includegraphics[height=2cm,width=0.9\columnwidth]{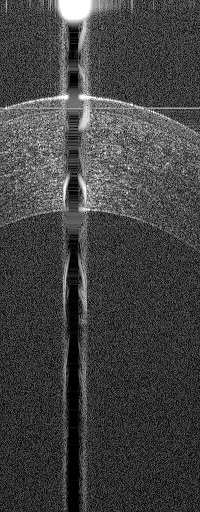}\\
\centerline{(a)}
\label{fig:d26_i0239_original}
\end{subfigure}
\begin{subfigure}[b]{0.29\columnwidth}
\centering
\includegraphics[height=2cm,width=0.9\columnwidth]{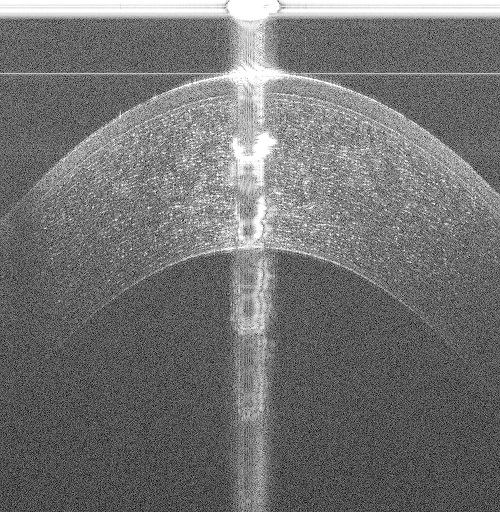}\\
\centerline{(b)}
\label{fig:d10_i35_CorNet}
\end{subfigure}
\begin{subfigure}[b]{0.125\columnwidth}
\centering
\includegraphics[height=2cm,width=0.9\columnwidth]{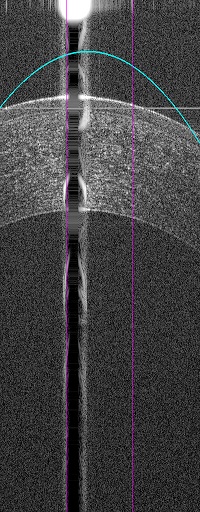}\\
\centerline{(c)}
\label{fig:d26_i0239_splitCols_prevMethodRes}
\end{subfigure}
\begin{subfigure}[b]{0.125\columnwidth}
\centering
\includegraphics[height=2cm,width=0.9\columnwidth]{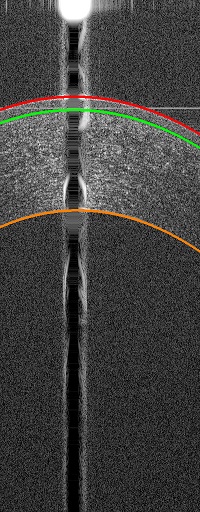}\\
\centerline{(d)}
\label{fig:d26_i0239_CorNet}
\end{subfigure}
\begin{subfigure}[b]{0.29\columnwidth}
\centering
\includegraphics[height=2cm,width=0.9\columnwidth]{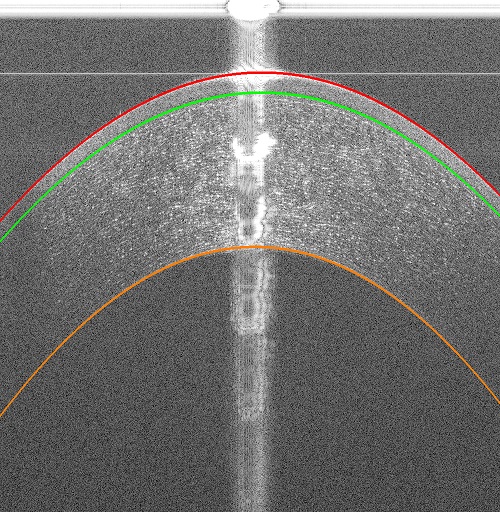}\\
\centerline{(e)}
\label{fig:d10_i35_CorNet}
\end{subfigure}
\caption{(a)-(b) Original B-scans from a 3$\times$3mm UHR-OCT and 6$\times$6mm SD-OCT volume; (c) Failed Epithelium segmentation result (cyan) from algorithms in \cite{LaRocca2011,Ge2012,Zhang2017}; (d)-(e) Our segmentation results for Epithelium (red), Bowman's layer (green), and Endothelium (orange) for images in (a) and (b).}
\label{fig:fig1}
\end{figure}

While many non-proprietary image analysis-based corneal interface segmentation approaches exist \cite{LaRocca2011,Ge2012,Williams2016,Rabbani2016,Zhang2017} in literature, they do not generalize to volumes acquired from different OCT scanners. These approaches are ad-hoc with key parameters being chosen manually; for example in Fig. \ref{fig:fig1}, recent approaches \cite{LaRocca2011,Ge2012,Zhang2017}, developed for images (B-scans) acquired by a Spectral Domain OCT (SD-OCT) scanner scanning a 6$\times$6mm area, failed while segmenting the Epithelium (shallowest layer) in 3$\times$3mm volumes acquired by a Ultra High Resolution OCT (UHR-OCT) scanner. Assumptions on the central artifact location \cite{LaRocca2011,Ge2012,Williams2016,Rabbani2016,Zhang2017} break down when they are located in different regions of the image (see Fig. \ref{fig:fig1}(c)). As shown in Figs. \ref{fig:fig1}(a) to \ref{fig:fig1}(c), a segmentation approach must perform reliably across datasets acquired with different scan settings from different scanners, even in the presence of strong vertical and horizontal specular artifacts. 

In recent years, neural networks have been shown to be successful in segmenting retinal tissue interfaces \cite{Roy2017,Shah2018,Devalla2018,Apo2017,Sedai2018} with great accuracy. In this paper, we detail a Convolutional Neural Network (CNN) based framework aimed at segmenting corneal interfaces. Our corneal interface segmentation network (CorNet) is purely data-driven, and learns to segment interfaces from examples drawn from different datasets acquired with different scanners. In contrast to current state-of-the-art approaches \cite{LaRocca2011,Zhang2017,Roy2017,Apo2017}, we show that our approach generalizes with better performance.

\noindent
\textbf{Contributions.} – 1) To the best of our knowledge, this is the first deep learning based approach to segment three corneal tissue interfaces. 2) We are the first to test a neural network on corneal datasets acquired with different scan settings from different OCT scanners. 3) We demonstrate the reliability of the approach through extensive validation on data acquired from different OCT scanners, and we establish superior performance over current state-of-the-art approaches. 4) We also investigate the performance of different downsampling and upsampling methods in our network, which are commonly used in segmentation tasks.

\section{METHODS}
\label{sec:methods}

\begin{figure}[!h]
\centering
\includegraphics[height=1.75cm,width=0.95\columnwidth]{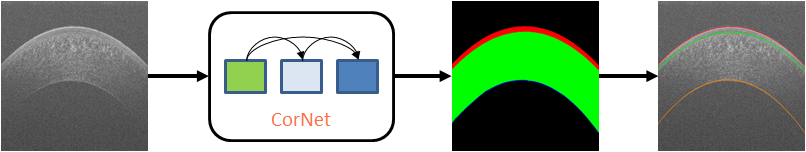}
\caption{Our framework takes as input an OCT image, predicts the location of corneal interfaces using the CorNet architecture, and fits curves to the detected interfaces.}
\label{fig:modelFlow}
\end{figure}

In this section, we outline the proposed CNN-based framework in Fig. \ref{fig:modelFlow} that segments three corneal interfaces.

\noindent
\textbf{Problem Statement.} Given a corneal OCT image $\mathcal{I}$, the task is to find a function $\mathcal{F : I \rightarrow L}$ that maps every pixel in $\mathcal{I}$ to a label $\mathcal{L} \in \{0,1,2,3\}$. Similar to \cite{LaRocca2011,Zhang2017}, the corneal interfaces to be segmented are: (1) Epithelium, (2) Bowman's Layer, and (3) Endothelium, with 0 being the background. 

%
\noindent
\textbf{Network Architecture.} Fully convolutional networks, such as the UNET \cite{Roy2017,Ronneberger2015} and BRUNET \cite{Apo2017}, are the state-of-the-art in retinal OCT segmentation. Such networks comprise of contracting and expanding branches, providing a dense output where each pixel is assigned the tissue class that it belongs to. The BRUNET architecture \cite{Apo2017} overcame problems of the UNET, such as holes in the segmentation, by modifying the UNET architecture. First, dilated convolutions \cite{Koltun2016,Devalla2018,Szegedy2015} were used in Inception-like blocks \cite{Szegedy2015} to increase the receptive field of each layer. Next, batch normalization \cite{Ioffe2015}, residual \cite{He2016} and bottleneck connections \cite{Szegedy2015}, and a feature map growth rate governed by a Fibonnaci sequence were incorporated. Finally, the input image was appropriately downsampled and connected to each layer. These changes greatly improved segmentation accuracy \cite{Apo2017} over the UNET. 

However, when applied to corneal OCT images, the BRUNET under-segmented poorly defined corneal interfaces, which are very common in anterior segment OCT imaging. As seen in Figs. \ref{fig:fig1} and \ref{fig:modelFlow}, these boundaries are corrupted by speckle noise, and have low signal-to-noise ratio (SNR). We empirically observed higher false positives in the final segmentation; one explanation is that discriminative features related to these boundaries being learned in earlier layers are lost through the network, and residual connections are unable to recover this information. 

One way to combine both coarse and fine image details is through the use of dense connections, which have been used to improve segmentation accuracy by encouraging heavy feature reuse through deep supervision \cite{Sedai2018,Huang2017,Jegou2017}. With dense connections, each layer is connected to all its preceding layers by feature map concatenation, allowing discernible features of faint boundaries to be retrieved across multiple scales. But, this comes at a cost of increased computation \cite{Jegou2017,Khosravan2018}, and we empirically determined that a densely connected network at a depth of 6 levels provides a good balance between segmentation accuracy and computational efficiency \cite{Apo2017,Khosravan2018}. Additionally, max pooling was better at maintaining features of interest through the network over average pooling and convolutions of stride 2 \cite{Khosravan2018}. Furthermore, nearest neighbor interpolation based upsampling followed by 3$\times$3 convolution \cite{Odena2016} performed better than bilinear interpolation based upsampling, bilinear interpolation + 3$\times$3 convolution \cite{Odena2016}, unpooling \cite{Roy2017,Noh2015}, and fractionally-strided convolutions \cite{Long2015}. 

\begin{figure}[!h]
\centering
\includegraphics[height=6.5cm,width=0.99\columnwidth]{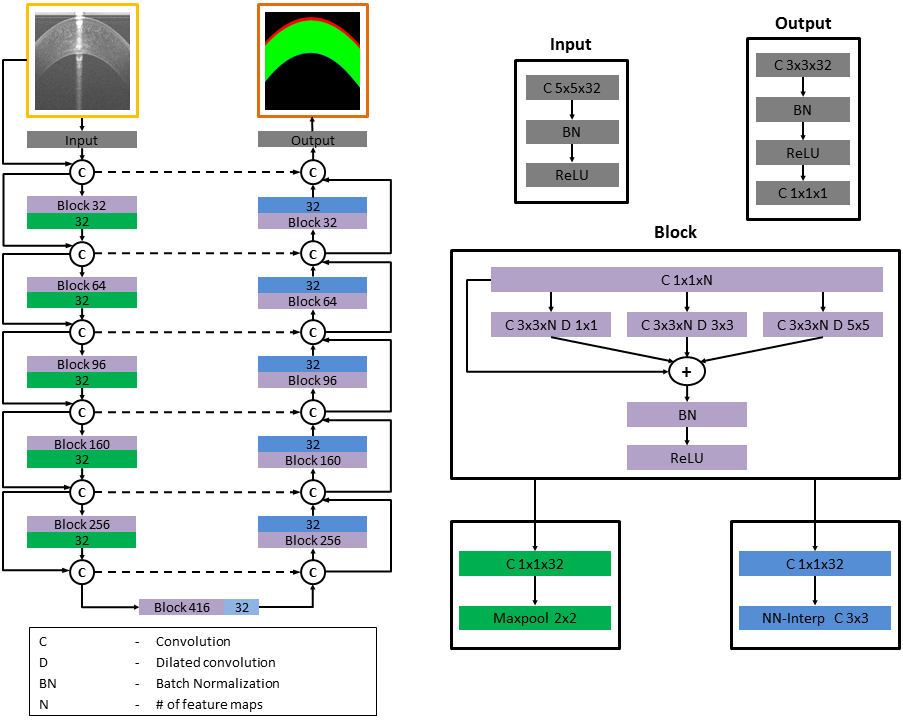}
\caption{Our network architecture comprises of contracting and expanding branches. The dark green and blue blocks represent downsampling and upsampling computations respectively. Our network makes efficient use of residual and dense connections to generate the corneal interface segmentation in the final image, where each pixel is assigned the label of the tissue it belongs to. The input image is split width-wise into a set of slices of dimensions 256$\times$1024 pixels, the network predicts an output for each slice, and the slices are aligned to recreate the original input dimension. Dense connections concatenate feature maps from previous layers. The light blue block at the bottom of the "U" does not perform upsampling, but it functions as a bottleneck and generates feature maps of the same dimensions as the output feature maps from the previous layer.}
\label{fig:arch}
\end{figure}

In our experiments, we adopted the BRUNET architecture \cite{Apo2017} as the base, and modified it based on our observations as shown in Fig. \ref{fig:arch}. Similar to \cite{Apo2017}, the number of output feature maps in each layer increased according to a capped Fibonacci sequence \{32,64,96,160,256,416\}, and limit the bottleneck feature map output to 32 to prevent feature map explosion. 

\noindent
\textit{Key modifications} to the architecture, which we incorporated were: 1) Dense connections were used to improve gradient information flow and prevent over-fitting; 2) Max pooling was used to pick the most discriminative features at the end of each downsampling layer; 3) Nearest neighbor interpolation + 3$\times$3 convolution was used to upsample feature maps in the expanding branch of the network. We name our corneal tissue interface segmentation architecture as \textit{CorNet}. 

\section{EXPERIMENTS AND RESULTS}
\label{sec:experiments}


\noindent
\textbf{Data.} De-identified datasets that had been previously acquired for an existing research database was used \cite{Mathai2018}. 48 volumes from both eyes of 8 subjects were acquired with different scan sizes using two OCT scanners; a Bioptigen SD-OCT scanner (Device 1) \cite{Wang2014}, and a high-speed ultra-high resolution OCT (hsUHR-OCT) scanner (Device 2) \cite{Srinivasan2006}. Device 1 had a 3.4\SI{}{\micro\meter} axial and 6\SI{}{\micro\meter} lateral spacing when scanning a 6$\times$6mm area, generating volumes of dimensions 1000$\times$1024$\times$50 (W$\times$H$\times$B-scans) pixels. Device 2 had a 1.3\SI{}{\micro\meter} axial and a 15\SI{}{\micro\meter} lateral spacing when scanning a 6$\times$6mm area, and a 7.5\SI{}{\micro\meter} lateral spacing when scanning a 3$\times$3mm area respectively, yielding volumes of size 400$\times$1024$\times$50 pixels. Each dataset was annotated by an expert grader (Grader 1) and a trained grader (Grader 2).

\noindent
\textbf{Setup.} Of the 48 datasets, 18 datasets were chosen for training, such that it contained a balanced number of datasets from both devices, i.e., six 6$\times$6mm datasets each from Device 1 and 2, and six 3$\times$3mm datasets from Device 2. The testing dataset comprised of 30 datasets; ten 6$\times$6mm datasets each from Device 1 and 2, and ten 3$\times$3mm datasets from Device 2. 5-fold cross-validation was conducted, and the model from the fold with the lowest validation loss was chosen for testing.

\begin{figure}[!h]
\centering
\begin{subfigure}[b]{0.125\columnwidth}
\vspace*{\fill}
  \centering
  \includegraphics[width=0.95\columnwidth,height=1.55cm]{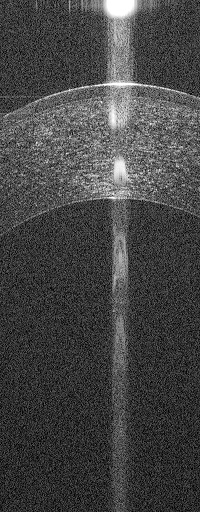}
  \centerline{(a)}\par\vfill
  \includegraphics[width=0.95\columnwidth,height=1.55cm]{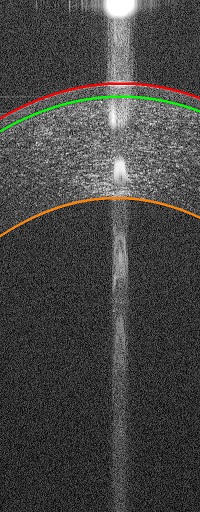}
  \centerline{(b)}
  \label{fig:cmodeResult}
\end{subfigure}\hfill
\begin{subfigure}[b]{0.125\columnwidth}
  \centering
  \includegraphics[width=0.95\columnwidth,height=1.55cm]{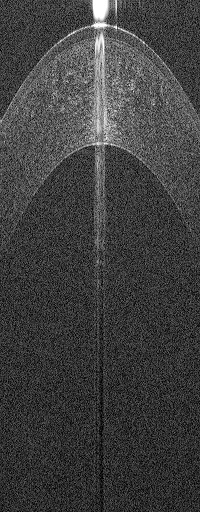}
  \centerline{(c)}\par\vfill
  \includegraphics[width=0.95\columnwidth,height=1.55cm]{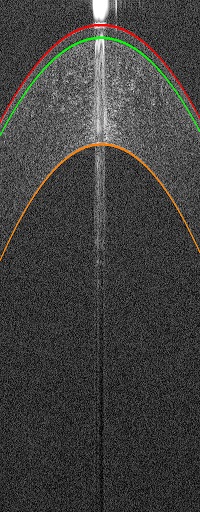}
  \centerline{(d)}
  \label{fig:cmodeResult}
\end{subfigure}\hfill
\begin{subfigure}[b]{0.25\columnwidth}
  \centering
  \includegraphics[width=0.9\columnwidth,height=1.55cm]{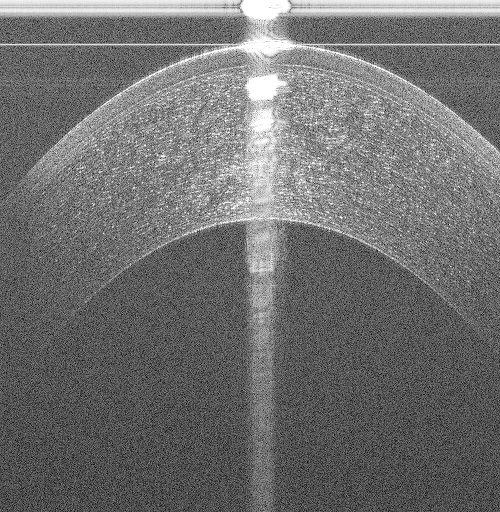}
  \centerline{(e)}\par\vfill
  \includegraphics[width=0.9\columnwidth,height=1.55cm]{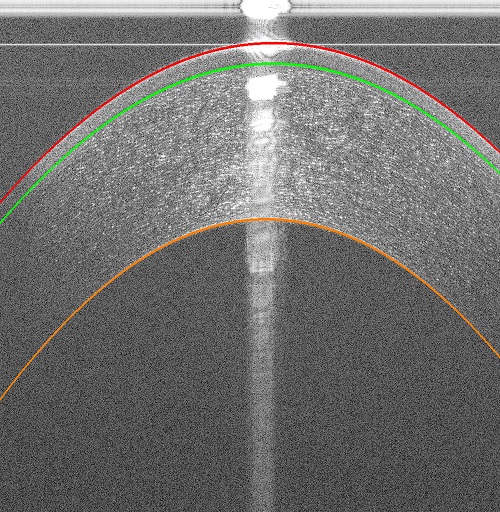}
  \centerline{(f)}
  \label{fig:cmodeResult}
\end{subfigure}\hfill
\begin{subfigure}[b]{0.25\columnwidth}
  \centering
  \includegraphics[width=0.9\columnwidth,height=1.55cm]{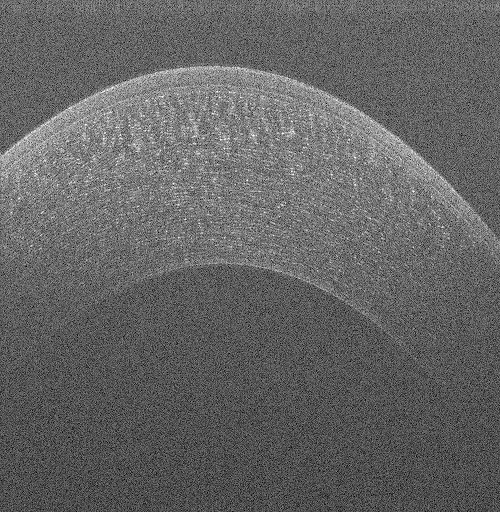}
  \centerline{(g)}\par\vfill
  \includegraphics[width=0.9\columnwidth,height=1.55cm]{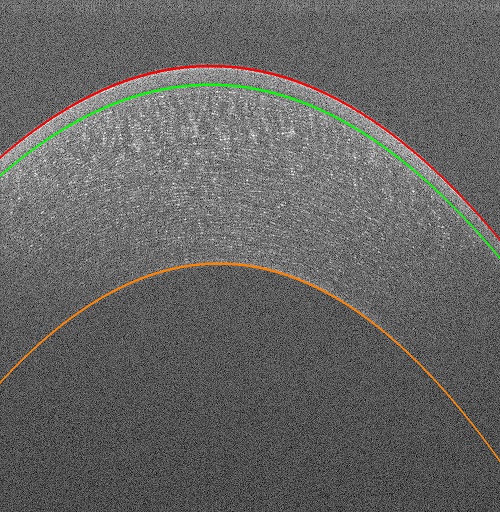}
  \centerline{(h)}
  \label{fig:cmodeResult}
\end{subfigure}
\caption{Original B-scans and segmented interfaces from different datasets:  (a)-(b) 3$\times$3mm UHR-OCT, (c)-(d) 6$\times$6mm UHR-OCT, and (e)-(h) 6$\times$6mm SD-OCT.}
\label{fig:res}
\end{figure}

\noindent
\textbf{Training.} Training a CorNet model with full-width OCT images is limited by available RAM on the GPU and by the varying image sizes obtained from OCT scanners. To address these issues, the input images were sliced width-wise \cite{Roy2017} into a set of images of dimensions 256$\times$1024 pixels, thereby preserving the OCT image resolution. Data augmentation \cite{Patrice2003} is done through horizontal flips, gamma adjustment, Gaussian noise addition, Gaussian blurring, Median blurring, Bilateral blurring, cropping, affine transformations, and elastic deformations. Similar to \cite{Apo2017}, the loss function used was Mean Squared Error (MSE), and the network was trained using the ADAM optimizer \cite{Kingma2015}. The batch size was set to 2. The learning rate was set to $10^{-3}$, and it was decreased by a factor of 2 if the loss did not improve for 5 epochs. Validation data comprised of 10\% of the training data, and the network was trained until the loss did not improve for 10 epochs, at which point we executed early stopping. The network with the lowest validation loss among all the folds was chosen for evaluation on the testing set. The prediction for each interface was then fitted with a curve \cite{LaRocca2011,Zhang2017,Mathai2018,Lowess1981} (see Fig. \ref{fig:res}).

\begin{figure}[!h]
\begin{subfigure}[b]{0.5\columnwidth}
\centering
\includegraphics[height=3.45cm,width=0.9\columnwidth]{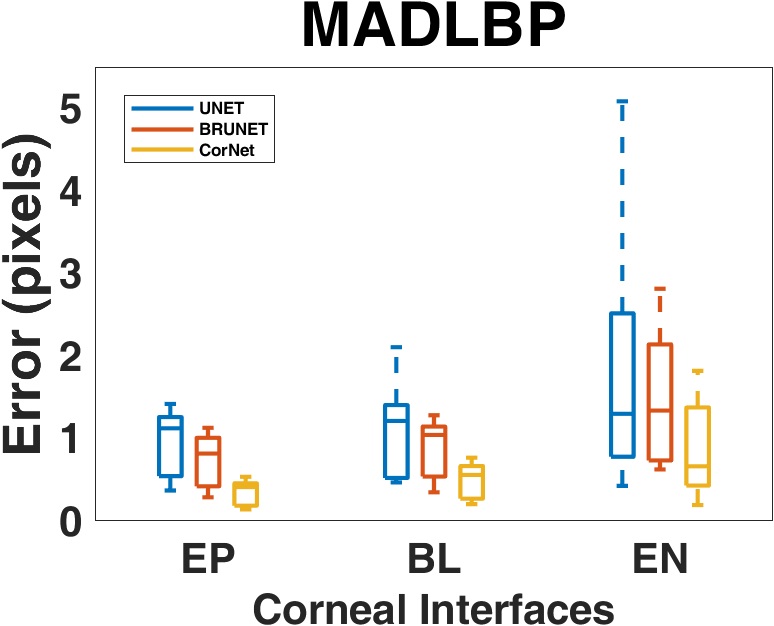}\\
\centering{(a)}
\end{subfigure}\hfill
\begin{subfigure}[b]{0.5\columnwidth}
\centering
\includegraphics[height=3.45cm,width=0.9\columnwidth]{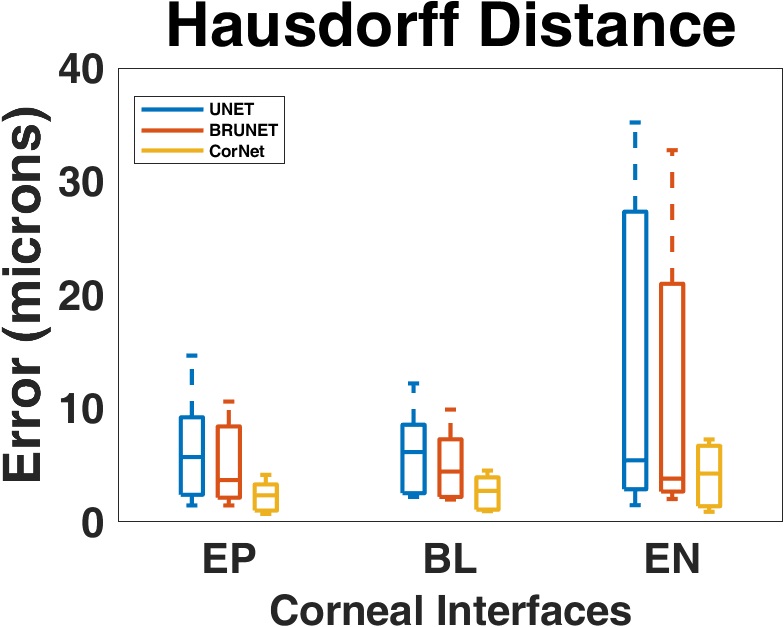}\\
\centering{(b)}
\end{subfigure}
\caption{Error comparison between expert annotation and automated segmentation (fitted with curves) obtained from different deep learning based methods across all 30 testing datasets.}
\label{fig:comparisonDL}
\end{figure}

\noindent
\textbf{Baseline Comparisons.} We extensively validated the performance of our CorNet architecture; first, we compared our results against those from the UNET \cite{Roy2017,Ronneberger2015} and BRUNET \cite{Apo2017} architectures as shown in Fig \ref{fig:comparisonDL}. Next, we compared our results against those obtained from \cite{LaRocca2011,Zhang2017} in Table \ref{table:traditionalComparison}; only 6$\times$6mm datasets from Device 1 were used as \cite{LaRocca2011,Zhang2017} solely considered datasets of this dimension. Finally, in Tables \ref{table:MADLBP} and \ref{table:HD}, we compared our results against each grader, and also computed the inter-grader variability measures to quantify our deviation from the agreement in ground truth between graders.

\noindent
\textbf{Metrics.} We computed the following metrics: 1) Mean Absolute Difference in Layer Boundary Position (MADLBP) and 2) Hausdorff Distance (HD) between the fitted curves. For consistency in comparison, we computed MADLBP as it was the metric (in pixels) of choice in \cite{LaRocca2011,Zhang2017}. However, MADLBP (Eq. \ref{eq1}) does not accurately quantify the distance error in microns between a particular pair of interfaces, which the Hausdorff distance (Eq. \ref{eq2}) captures instead. Dice similarity did not provide error in microns, and thus was not computed in this work. Metrics were computed for the Epithelium (EP), Bowman's Layer (BL), and Endothelium (EN). In Eqs. \ref{eq1} and \ref{eq2}, $G$ and $S$ are the set of points in the ground truth annotation and segmentation (fitted with curves) respectively. $y_{G}(w)$ is the mean Y-coordinate (rounded down) of the points in $G$ whose X-coordinate is $w$, and similarly for $y_{S}(w)$. $d_{S}(x)$ is the distance of a point $x$ in $G$ to the closest point in $S$, and similarly for $d_{G}(x)$. 
\begin{align}
  	\textnormal{MADLBP} &= \frac{1}{W} \sum\limits^{W-1}_{w=0} \abs{y_{G}(w) - y_{S}(w)} \label{eq1}\\
	\textnormal{HD} &= \max\bigg(\underset{x \in G}{\max} \ d_{S}(x), \ \underset{x \in S}{\max} \ d_{G}(x) \bigg) \label{eq2}
\end{align}
\begin{table}[!ht]
\centering\fontsize{9}{11}\selectfont
\caption{Comparison of Mean Absolute Difference in Layer Boundary Position (MADLBP) error between traditional methods against the proposed deep learning based approach on ten 6$\times$6mm volumes from Device 1. Only expert annotations were used for comparison. Errors are in pixels.}
\begin{tabular}{cccc}\hline
\toprule
Approach					    & EP                            & BL                            & EN \\ 	\midrule

LaRocca et al. \cite{LaRocca2011}				& 0.84 $\pm$ 0.31           & 1.12 $\pm$ 0.4                & 1.97 $\pm$ 2.26   \\
Zhang et al. \cite{Zhang2017}				& 0.69 $\pm$ 0.24               & 0.91 $\pm$ 0.35               & 1.73 $\pm$ 1.98   \\
Proposed				    & \textbf{0.33 $\pm$ 0.21}      & \textbf{0.42 $\pm$ 0.13}      & \textbf{0.79 $\pm$ 0.19}   \\ \bottomrule
\end{tabular}
\label{table:traditionalComparison}
\end{table}
\begin{table}[!ht]
\centering\fontsize{9}{11}\selectfont
\caption{Mean Absolute Difference in Layer Boundary Position (MADLBP) error across 6$\times$6mm datasets from Device 1 (top half), and 3$\times$3mm and 6$\times$6mm datasets from Device 2 (bottom half). Errors are in pixels.}
\begin{tabular}{cccc}\hline
\toprule
Layer       & Grader 1                          & Grader 2                          & Inter-Grader \\ 	\midrule

EP          & 0.33 $\pm$ 0.21                   & 0.41 $\pm$ 0.14                   & 0.49 $\pm$ 0.07   \\
BL          & 0.42 $\pm$ 0.13                   & 0.68 $\pm$ 0.17                   & 0.51 $\pm$ 0.06   \\
EN          & 0.79 $\pm$ 0.19                   & 0.84 $\pm$ 0.34                   & 0.56 $\pm$ 0.22   \\ \midrule

EP          & 0.32 $\pm$ 0.09                   & 0.49 $\pm$ 0.13                   & 0.49 $\pm$ 0.09   \\
BL          & 0.41 $\pm$ 0.13                   & 0.61 $\pm$ 0.15                   & 0.5 $\pm$ 0.09   \\
EN          & 0.93 $\pm$ 0.19                   & 1.45 $\pm$ 0.39                   & 0.61 $\pm$ 0.29   \\ \bottomrule
\end{tabular}
\label{table:MADLBP}
\end{table}
\begin{table}[!ht]
\centering\fontsize{9}{11}\selectfont
\caption{Mean Hausdorff Distance (HD) error across 6$\times$6mm datasets from Device 1 (top half), and 3$\times$3mm and 6$\times$6mm datasets from Device 2 (bottom half). Errors are in microns.}
\begin{tabular}{cccc}\hline
\toprule
Layer	& Grader 1                          & Grader 2                          & Inter-Grader      \\ 	\midrule

EP		& 3.17 $\pm$ 1.04                   & 4.46 $\pm$ 1.23                   & 3.21 $\pm$ 0.52   \\
BL		& 3.52 $\pm$ 1.39                   & 4.15 $\pm$ 1.05                   & 3.22 $\pm$ 0.5   \\
EN		& 5.55 $\pm$ 2.24                   & 6.7 $\pm$ 3.78                    & 4.05 $\pm$ 1.2   \\ \midrule

EP		& 1.52 $\pm$ 0.42                   & 1.63 $\pm$ 0.42                   & 1.21 $\pm$ 0.21   \\
BL		& 1.89 $\pm$ 0.62                   & 1.95 $\pm$ 0.68                   & 1.23 $\pm$ 0.22   \\
EN		& 3.05 $\pm$ 1.08                   & 4.03 $\pm$ 1.34                   & 1.76 $\pm$ 0.62   \\ \bottomrule
\end{tabular}
\label{table:HD}
\end{table}

\section{Discussion}
\label{sec:discussion}


From Fig. $\ref{fig:comparisonDL}$ and Table $\ref{table:traditionalComparison}$, our network outperformed the current deep learning \cite{Roy2017,Ronneberger2015,Apo2017} and traditional approaches \cite{LaRocca2011,Zhang2017} respectively. Paired t-tests conducted between our approach and every baseline established that for each metric our results were statistically significant (\textit{p} $<$ 0.05).

The MADLBP error (in pixels) and mean Hausdorff distance (in microns) across 6$\times$6mm datasets from Device 1 (Tables $\ref{table:MADLBP}$ and $\ref{table:HD}$, top halves) for the expert grader is slightly lower when contrasted against the trained grader. We attribute this to the diffuse appearance of corneal interfaces \cite{Kuo2012,LaRocca2011,Roy2017} and lower axial resolution of Device 1 (3.4\SI{}{\micro\meter}), thereby causing an expected deviation between the grader annotations, which is reflected in the inter-grader MADLBP error. Similar measures on the MADLBP error (in pixels) and mean Hausdorff distance (in microns) across 3$\times$3mm and 6$\times$6mm datasets from Device 2 (Tables $\ref{table:MADLBP}$ and $\ref{table:HD}$, bottom halves) were observed. Overall, we closely matched the inter-grader error across all datasets for the EP and BL interfaces, and in some cases, perform better than the agreement between graders.

With respect to the EN, our errors were worse than the inter-grader agreement on the interface location. We attribute this to the low SNR in many corneal images, particularly at the left and right edges of the EN where the signal dropoff is substantial \cite{LaRocca2011}. In these regions, the graders mentally extrapolated their annotations for this interface with poorly defined boundaries, which were usually obfuscated by speckle noise. When a curve is fitted to both the annotation and prediction, there is a small degree of error during the comparison, which is unavoidable. This behavior has also been observed in \cite{LaRocca2011, Zhang2017}. However, our EN errors were considerably better than the measured MADLBP and HD errors for the state-of-the-art image analysis-based and deep learning based approaches. The CorNet took $\sim$15.1 s (Python) to segment an entire volume of 50 images of dimensions 1000$\times$1024 pixels, at $\sim$302 ms per image. This is in contrast to 56.5 s for \cite{LaRocca2011} (Matlab), $\sim$26.1 s for \cite{Zhang2017} (Matlab), $\sim$6.25 s for BRUNET (Python), and $\sim$10.75 s for UNET (Python); CorNet is slower than UNET or BRUNET due to dense connections. The results were calculated on a desktop using a 3.10 GHz Intel Xeon processor, 64 GB RAM, and a NVIDIA Titan Xp GPU.

\noindent
\textbf{Major Observations.} 1) The proposed CorNet architecture consistently outperforms the state-of-the-art image analysis-based and deep learning-based approaches for the task of corneal tissue interface segmentation. 2) Maxpooling is optimal for feature selection across the common downsampling choices. 3) Nearest neighbor interpolation based feature map upsampling followed by 3$\times$3 convolution improved segmentation over other upsampling operations. 4) Dense connections increased segmentation accuracy due to greater gradient information flow through the network.

\section{CONCLUSION AND FUTURE WORK}
\label{sec:conclusion}

To the best of our knowledge, we have presented the first CNN-based framework to segment three corneal tissue interfaces in datasets that have been acquired from different OCT scanners with different scan settings. Our CorNet results have been extensively validated against the annotations of two graders, current state-of-the-art approaches in deep learning, and against traditional approaches towards corneal interface segmentation. Future work is aimed at extending our work to pathological corneas, and using the segmentation to drive the registration of B-scans with out-of-plane tissue motion. 

\noindent
\textbf{Acknowledgements.} We thank our funding sources: NIH 1R01EY021641, Core Grant for Vision Research EY008098-28. We thank NVIDIA Corporation for their GPU donations. We also thank Haewon Jeong, Wonmin Byeon, Bo Wang, Katie Lucey, and Gadi Wollstein for helpful comments.


\end{document}